\documentclass[3p,12pt]{elsarticle}
\usepackage[numbers]{natbib}          
\usepackage{graphicx}
\usepackage{amsmath,amssymb}
\usepackage{booktabs}
\usepackage{enumitem}
\usepackage[hidelinks]{hyperref}
\usepackage{orcidlink}
\usepackage{listings}   
\usepackage[most]{tcolorbox}
\usepackage[table]{xcolor} 
\usepackage{float}
\usepackage{setspace}
\usepackage{lineno}
\usepackage{float}
\usepackage{multirow}
\usepackage{csquotes}
\usepackage{times}
\newcommand{\encoder}{\ensuremath{\mathcal{E}}}

\usepackage{etoolbox}
\usepackage{xurl}
\usepackage{IEEEtrantools}

\usepackage{amssymb}

\usepackage{amsmath}


\makeatletter
\def\ps@pprintTitle{%
 \let\@oddhead\@empty
 \let\@evenhead\@empty
 \def\@oddfoot{\reset@font\hfil} 
 \let\@evenfoot\@oddfoot
}
\makeatother

\usepackage{caption}
\begin{document}
\bstctlcite{bstctl:nodash}
\doublespacing %
\makeatletter
\def\bibfont{\normalsize}          
\appto{\bibsection}{\doublespacing} 
\makeatother
\begin{frontmatter}

\title{\textbf{Scene Graph-Guided Generative AI Framework for Synthesizing and Evaluating Industrial Hazard Scenarios}}

\author[1]{Sanjay Acharjee}
\author[2]{Abir Khan Ratul}
\author[3]{Diego Patiño, Ph.D.}
\author[4]{Md Nazmus Sakib, Ph.D.\corref{cor1}} 

\address[1]{Ph.D. Student, Dept. of Civil Eng., University of Texas at Arlington. E-mail: sanjay.acharjee@uta.edu}
\address[2]{Ph.D. Student, Dept. of Civil Eng., University of Texas at Arlington. E-mail: abirkhan.ratul@uta.edu}
\address[3]{Assistant Professor, Dept. of Computer Sci. \& Eng., University of Texas at Arlington. E-mail: diego.patino@uta.edu}
\address[4]{Assistant Professor, Dept. of Civil Eng., University of Texas at Arlington. E-mail: mdnazmus.sakib@uta.edu}
\cortext[cor1]{Corresponding author}

\begin{abstract}
Training vision models to detect workplace hazards accurately requires realistic images of unsafe conditions that could lead to accidents. However, acquiring such datasets is difficult because capturing accident-triggering scenarios as they occur is nearly impossible. To overcome this limitation, this study presents a novel scene graph-guided generative AI framework that synthesizes photorealistic images of hazardous scenarios grounded in historical Occupational Safety and Health Administration (OSHA) accident reports. OSHA narratives are analyzed using GPT-4o to extract structured hazard reasoning, which is converted into object-level scene graphs capturing spatial and contextual relationships essential for understanding risk. These graphs guide a text-to-image diffusion model to generate compositionally accurate hazard scenes. To evaluate the realism and semantic fidelity of the generated data, a visual question answering (VQA) framework is introduced. Across four state-of-the-art generative models, the proposed VQA Graph Score outperforms CLIP and BLIP metrics based on entropy-based validation, confirming its higher discriminative sensitivity.

\end{abstract}




\end{frontmatter}



\section{Introduction}\label{sec:intro}

\noindent The modern industrial landscape is undergoing a profound transformation, driven by advancements in automation, robotics, and artificial intelligence (AI)~\cite{osha_analysis_robot}. This new era promises unprecedented efficiency, but also introduces a parallel challenge with the rise of new, fast-changing workplace hazards that traditional safety systems were never built to handle. As intelligent systems and human workers increasingly collaborate in this dynamic environment, the very nature of risk is evolving, creating a critical safety lag in which traditional protocols are fundamentally mismatched with modern workplace realities. These challenges are supported by alarming real-world safety statistics, such as approximately 2.6 million occupational injuries and illnesses reported in the U.S. in 2023~\cite {labor_stat_23_1}. A worker dying every 96 minutes due to work-related incidents shows the severity of these injuries~\cite{labor_stat_22,labor_stat_23_2}.

This issue is widening because traditional safety protocols, designed for a more static industrial era, are fundamentally reactive. They struggle to keep pace with the dynamic nature of modern work, where the introduction of new technologies like robotics and automation creates novel hazards that safety regulations have not yet addressed. To manage these emerging risks, a shift toward proactive, data-driven safety systems powered by AI is essential. However, the development of such intelligent systems is always slowed by data shortage. Resources like the Occupational Safety and Health Administration’s (OSHA) Severe Injury Reports (SIR)~\cite{OSHA_SevereInjury} offer detailed textual descriptions of accidents but lack corresponding critical visual data. This lack of visual data creates a significant issue, limiting the ability to train intelligent computer vision models needed for a modern, automated safety framework capable of reasoning about underlying hazards in complex environments, rather than just detecting routine safety issues like PPE (Personal Protective Equipment) compliance.

The challenge of identifying subtle or context-specific hazards requires more than just object detection; it demands a deeper, causal understanding of a scene. Modern Vision-Language Models (VLMs) are powerful tools uniquely suited to this task, as they combine visual data processing with the sophisticated reasoning and pattern-recognition capabilities of large language models~\cite{vaswani2017attention}. These models can understand not just what objects are in a scene, but also the critical relationships and contextual elements that define it. However, their application in workplace safety is limited by the lack of appropriate validation and training data. Datasets that exist often lack the specific structure needed to train models to do predictive reasoning as they fail to capture the environmental factors that contribute to an incident~\cite{al2024harnessing}. The core of the problem is that obtaining real-world visual data capturing the precise conditions just before an accident is practically impossible. These events are sudden, unforeseen, and rarely documented, creating an unbridgeable gap in the training data needed to build AI systems that can proactively prevent accidents before they occur~\cite {osha_analysis1}.

To overcome this critical data scarcity, a novel approach is proposed that systematically translates historical accident reports into rich, synthetic visual data. Instead of attempting to capture real-world pre-accident scenarios, the framework leverages OSHA’s accident records to generate realistic, context-aware visual representations of hazards. By utilizing sophisticated generative AI models, such as large language models like GPT -4~\cite{gpt4} and diffusion-based models~\cite{stable_diffusion}, a scalable and diverse dataset for reasoning over historical reports can be created, enabling high-fidelity image synthesis. This method effectively bridges the gap between textual knowledge and the visual data required to train the next generation of proactive safety AI.

However, the synthesis of visually realistic images solves only a part of the problem. For synthetic data to be useful in training safety models, the images must accurately represent real-life hazardous situations. AI-generated scenes often contain subtle errors, such as physically unrealistic object interactions or inconsistent spatial relationships, which reduce their reliability. Standard evaluation metrics often overlook these issues, as they typically measure overall prompt alignment rather than the compositional accuracy needed to confirm whether a scene is physically valid and contextually appropriate for safety analysis.. This creates a critical need for a more rigorous evaluation framework to validate the fidelity of synthetically generated hazardous conditions.

This work presents a complete, end-to-end framework for generating and evaluating synthetic workplace hazard scenarios, providing a much-needed resource for training proactive safety models. Our primary contributions are threefold: (1) a novel methodology that uses scene graphs as an intermediate representation to guide the generation of semantically accurate and context-aware images from OSHA accident narratives; (2) A novel VQA-based evaluation metric that proves more sensitive and reliable for assessing semantic fidelity than standard benchmarks; (3) A complete, scalable dataset with generation code and evaluation tools to support the development of more robust safety systems across industries.

\section{Related Work}\label{sec:related}
\subsection{Generative AI for Hazard Synthesis}
AI-based systems for detecting workplace hazards have advanced considerably in recent years. However, these systems fail to recognize subtle or context-specific risks because they focus on individual objects rather than the complex spatial and environmental relationships that define a hazardous scenario. A significant limitation in training and validating these systems is the scarcity of labeled visual datasets that correspond to real-world accidents~\cite{11_C}. Although many industries maintain detailed text-based accident reports that contain valuable information about incident mechanisms, the lack of corresponding visual data remains a major bottleneck for training robust computer vision models~\cite{kim2024image}. Recent advancements in generative AI, particularly Generative Adversarial Networks (GANs)~\cite{goodfellow2020generative} and diffusion-based models~\cite{stable_diffusion}, have opened a promising avenue to address this data gap. This has led to the development of generative AI-driven data augmentation and generation frameworks that generate scarce real-world data using synthetic images~\cite{lee2025generative}, enabling the creation of large-scale, diverse datasets representing rare or high-risk scenarios that would be impractical to capture in reality~\cite{lee2025generative,nath2020deep}.

The efficacy of these methods depends on the generative model's ability to produce visually plausible images and compositionally coherent scenarios. For instance, workplace risk arises not from a single object but from spatial conditions and interactions among multiple entities. However, many current generative methods fall short in this regard, often creating images focused on isolated objects~\cite{kim2024image} while failing to capture the complex interactions critical for a valid risk assessment. This limitation is a well-documented challenge in the broader field of text-to-image generation, where even state-of-the-art models struggle with compositional accuracy, as highlighted by benchmarks like T2I-CompBench~\cite{compbench}. To overcome these gaps, researchers are moving toward integrating structured reasoning capabilities into the generative process ~\cite{qwen_image_edit}. By fundamentally changing how the model understands the scene it is generating, rather than just using robust models such as Flux~\cite{flux} or NVIDIA Cosmos~\cite{cosmos}. Scene graphs have emerged as a particularly effective framework for this purpose, as they encode objects as nodes and their spatial or semantic relationships as edges, offering a structured representation of the underlying scene logic. The foundational work on generating images from scene graphs provides a clear path for improving existing image generation pipelines, demonstrating that explicit relational structure is key to generating coherent, contextually accurate images~\cite{johnson2015image}. By adopting such structured approaches, generative AI can evolve from simply creating images of hazards to synthesizing entire, physically plausible hazardous scenarios, enabling models to move beyond routine tasks such as real-time PPE detection toward genuine hazard reasoning and contextual risk assessment.~\cite{ayata2024ai}.

\subsection{Limitations in Evaluation Metrics}
Despite rapid progress in generative AI for hazard synthesis, previous studies have largely overlooked rigorous evaluation methodologies. Most existing works rely on generic image quality or text–image alignment metrics, which are inadequate for assessing whether a generated scene accurately represents a hazardous situation. Early automatic evaluation metrics for text-to-image (T2I) generation lean on single embedding-based scores that estimate how well an image matches its prompt without any human-annotated references. The best known metric is perhaps the Contrastive Language-Image Pre-training Score (CLIPScore)~\cite{clip_paper}. CLIPScore operates by computing the cosine similarity between the text embedding vector $u$ (obtained from the prompt) and the image embedding vector $v$ (obtained from the generated image), as follows:
\begin{equation}
\operatorname{cos\_sim}(u, v)= \frac{\langle u,\, v\rangle}{||u||\;||v||} \label{eq:cosine_sim}
\end{equation}
Here, $\langle u, v\rangle$ denotes the dot product between the text embedding vector $u$ and the image embedding vector $v$, while $||u||$ and $||v||$ represent their respective Euclidean norms. The CLIP Score was initially designed to track human judgments in captioning tasks and was later widely used for text-to-image generation~\cite{clip_paper,dalle}. BLIPScore~\cite{blip} is another quality metric for image generation that follows the same principles as CLIPScore. However, BLIPScore uses the image and text matching head or embeddings from BLIP~\cite{blip} or BLIP 2~\cite{blip2} to evaluate the semantic alignment between the generated image and its corresponding textual description. It has become a common, reference-free measure for assessing how well a generated image matches its prompt in recent evaluations and benchmarks. These metrics are popular because they are simple, fast, and model agnostic. However, they have an important limitation in how they function. By converting the entire image and its text prompt into single feature vectors in a shared latent space, they capture only overall similarity instead of detailed compositional accuracy. As a result, they can judge whether an image fits the general theme of a prompt but fail to recognize specific spatial relationships, object interactions, or physical inconsistencies, all of which are essential for evaluating hazard scenarios.~\cite{compbench, clip_paper, dalle_eval}.

On the other hand, metrics such as the Fréchet Inception Distance (FID)~\cite{fid} compare the feature distributions of generated and real images to measure realism and diversity. The process begins by using a pre-trained InceptionV3 model~\cite{inceptionv3} to extract high-level feature vectors from both the set of real images and the set of generated images, creating two distinct clouds of data points in a high-dimensional space. To compare these clouds efficiently, each is summarized as a multivariate Gaussian distribution, defined by a mean, $\mu$, representing the average features and a covariance, $\Sigma$, representing the diversity and spread of features. The FID score is then calculated as the Fréchet distance (FID) between these two distributions. This distance metric quantifies the difference between them by considering both the squared distance between their means and the difference in their covariance matrices,
\begin{equation}
\operatorname{FID} = \left\| \mu_{\text{real}} - \mu_{\text{gen}} \right\|_2^2 + \text{Tr}\left(\Sigma_{\text{real}} + \Sigma_{\text{gen}} - 2\left(\Sigma_{\text{real}}\Sigma_{\text{gen}}\right)^{1/2}\right) 
\end{equation}
Here, $\mu_{\text{real}}$ and $\mu_{\text{gen}}$ denote the mean feature vectors of the real and generated image sets, respectively, and 
$\Sigma_{\text{real}}$ and $\Sigma_{\text{gen}}$ represent their covariance matrices. $\mathrm{Tr}(\cdot)$ denotes the matrix trace, and $(\Sigma_{\text{real}}\Sigma_{\text{gen}})^{1/2}$ denotes the principal matrix square root. A lower $\operatorname{FID}$ value indicates that the generated images are more similar to real ones in both average features and overall diversity. A lower FID score indicates that the generated images are a better match for the real images in both their central tendency and their variety. However, FID completely ignores the text prompt used for generation. Consequently, it cannot verify whether specific objects, counts, or relationships match the description. Thus a model can achieve an excellent FID score while failing to adhere to key parts of the prompt.

Although metrics such as CLIPScore and FID offer convenient ways to assess image–text alignment or visual realism, they provide only a narrow view of model performance. Recent holistic evaluations have shown that text-to-image systems must be judged across multiple dimensions, not just a single score. A holistic benchmark that assessed $12$ aspects, including alignment, visual quality, originality, reasoning, bias, safety, fairness, and efficiency, found that no single model excelled across all dimensions~\cite{lee2023holistic}. More importantly, human judgments correlated very weakly with popular metrics, such as CLIPScore and FID, for key qualities in the benchmark, including photorealism and aesthetics. These findings highlight that relying on a single-number metric can mask important strengths and weaknesses, and that the choice of evaluation method can implicitly favor certain model behaviors~\cite{lee2023holistic}.

Building on these insights, recent work has begun exploring large language model (LLM)–based evaluation approaches that aim to better capture semantic and holistic understanding. A growing trend in text-to-image evaluation involves using LLMs to evaluate image quality and alignment from a prompt. One representative method is LLMScore~\cite{lu2023llmscore}, which generates overall and object-level image descriptions and then queries an LLM to assess how well the image corresponds to the prompt, producing both a quantitative score and a textual explanation. Studies using datasets such as DrawBench~\cite{drawbech} and PaintSkills~\cite{paintskill} have shown that LLMScore achieves stronger alignment with human ratings than CLIP or BLIP, especially for complex, object-level compositions. However, this approach relies on additional perception or captioning models and large-scale LLMs, introducing challenges related to cost, reproducibility, and evaluator bias. In fact, some studies report that even advanced LLM-based or question-answering metrics do not consistently outperform simpler feature-based methods like CLIPScore when evaluating complex real-world semantics~\cite{who_evals_evals}. These results suggest that while LLM-based metrics are promising, caution is needed when interpreting single leaderboard scores from any one automatic evaluation method.

Another approach, called TIFA~\cite{tifa}, treats visual question answering (VQA) as a means to execute the prompt. TIFA takes the text prompt and automatically generates specific yes or no questions, such as whether a certain object is present, its color is correct, or whether two objects have the right spatial relationship. It then uses a VQA model to answer those questions based on the generated image, producing a detailed and interpretable reliability score. This method not only captures fine-grained details but also clarifies where a model succeeded or failed. TIFA authors demonstrate stronger alignment with human judgments than CLIP-based similarity measures, while also revealing persistent weaknesses in areas such as counting and spatial reasoning across various text-to-image models. Other methods have also shown promising results using VQA models to evaluate and improve text-to-image alignment~\cite{singh2023divide}.

Although these approaches represent important progress, evaluating text-to-image systems remains challenging. Human evaluation is still considered the gold standard, yet it is often performed with inconsistent procedures, small sample sizes, and limited reporting ~\cite{human_eval_issue}. These problems lead to low agreement between annotators and unstable model rankings. The same work showed that widely used automatic metrics can be misleading. For example, the FID can rank systems incorrectly, and CLIPScore can saturate or be gamed. Statistical power analysis, a calculation used to determine the minimum sample size required to reliably detect a true effect, reveals the fragility of human evaluation when sample sizes are too small. Researchers have found that testing on fewer than about $100$ prompts, or having too few raters per image, renders the experiment statistically underpowered. This means the results are not robust enough to distinguish true model performance from random noise, which can cause the ranking of models to change arbitrarily, highlighting how fragile headline results can be without standardized processes and sufficient data~\cite{human_eval_issue}.

\subsection{Scene Graphs for Hazard Reasoning and Generation}
A proper understanding and description of complex visual scenes requires not only identifying objects, but also interpreting how those objects interact and relate to one another. Scene graphs have emerged as a powerful framework for this purpose, providing a structured representation that encodes objects as nodes and their relationships as directed edges~\cite{xu2017scene}. Early foundational research has demonstrated that scene graphs can serve as a rich intermediate layer for generating coherent and semantically consistent images, revealing a strong connection between the symbolic graph structure and the final visual output ~\cite{Image_Generation_from_Scene_Graphs}.

Building on this foundation, recent work has leveraged scene graphs for high-stakes reasoning tasks such as hazard detection~\cite{zhang2022automatic}. Their explicit and compositional structure makes them particularly well-suited for modeling the spatial and contextual conditions that give rise to safety risks. In applications like automated construction site monitoring, systems can now generate scene graphs directly from site imagery and analyze them to detect unsafe arrangements, hazardous interactions, or violations of safety protocols. This structured reasoning approach not only identifies hazards but also explains why a scene is unsafe in a way that is interpretable and actionable~\cite{zhang2022automatic}. Integrating these vision-generated graphs with domain knowledge from safety regulations further enhances their ability to perform hazard classification and reasoning with high accuracy~\cite{liu2023automatic,zhang2022automatic}. Beyond construction, similar methods have been applied in other safety-critical domains, including elderly-centric indoor monitoring~\cite{jiang2025hierarchical} and autonomous driving, where modeling pedestrian–vehicle interactions as scene graphs improves risk assessment~\cite{liu2023learning}.

\section{Methodology}\label{sec:method}
\noindent To generate realistic visual representations of workplace hazards, a methodology was developed that integrated large language models~\cite{vaswani2017attention}, generative diffusion models~\cite{stable_diffusion}, and vision–language evaluation techniques into a unified pipeline. The OSHA accident narratives were first analyzed using GPT-4o~\cite{gpt4} to extract structured hazard rationales and classify incidents. The text embeddings of these rationales were then calculated, clustered, and transformed into scene graphs using LLaMA 3~\cite{dubey2024llama}. Through this process, the spatial and contextual relationships among key objects were captured. The resulting scene graphs were subsequently used to guide the image generation process through text-to-image models such as Stable Diffusion~\cite{stable_diffusion}. In addition, the graphs supported a structured evaluation of the generated images through a graph-aligned vision–question-answering strategy.

\subsection{Classification and Clustering of OSHA Accident Narratives with GPT-4o Reasoning}\label{sec:osha}

The initial stage of the methodology involves classifying narrative reports from OSHA’s Severe Injury Reports (SIR) dataset using the reasoning capability of GPT-4o~\cite{llm_crash_report}. The narratives in these reports provide detailed descriptions of workplace incidents, offering critical insights into how injuries occur. The narratives typically include information such as the sequence of events leading to the injury, the type of accident (e.g., falls, struck-by incidents, caught-in-between), and the conditions that contributed to the injury. The reports are structured with a focus on the event's nature, including the body part affected and the source of the injury. This study included injury reports from January 2015 to January 2024, comprising a total of 89,068 entries. Each record contained details such as the event date, employer name, and a brief description of the incident. The dataset also included a final narrative that provided a more detailed explanation of the circumstances surrounding each injury, offering insights into how the incident occurred. Additionally, the dataset categorized injuries by affected body part, nature of the injury, and source of the incident. These narratives served as a rich source of information for analyzing patterns in workplace injuries. Table~\ref{tab:osha} presents a description of the available fields in the dataset.

\begin{table}[h!]
\fontsize{11pt}{13.2pt}\selectfont
\centering
\caption{Description of information contained in the OSHA Severe Injury Reports.}
\vspace{5pt}
\label{tab:osha}
\begin{tabular}{@{}p{4.25cm}p{9cm}@{}}
\toprule
\textbf{Column Name} & \textbf{Description} \\ 
\midrule
EventDate & The date when the injury event occurred. \\
Employer & The name of the employer where the injury occurred. \\
Hospitalized & Indicator (1 or 0) if the worker was hospitalized. \\
Amputation & Indicator (1 or 0) if the injury resulted in amputation. \\
Final Narrative & A detailed description of the accident. \\
NatureTitle & General classification of the injury's nature (e.g., fracture). \\
Part of Body Title & The body part affected by the injury (e.g., lower leg). \\
EventTitle & The type of event or accident (e.g., fall, malfunction). \\
SourceTitle & The source or cause of the injury (e.g., machinery, slip). \\
Secondary Source Title & Contributing factors or secondary sources. \\ 
\bottomrule
\end{tabular}
\end{table}

Each narrative was evaluated to determine whether it included an identifiable hazard and whether that hazard could have reasonably contributed to the incident. This classification step segmented the dataset into three mutually exclusive categories: preventable hazards, non-external factor incidents, and cases with insufficient information.

The preventable hazards (avoidable) category included incidents in which an identifiable and avoidable external hazard, such as an unguarded edge, misplaced tool, or spill, was clearly described as contributing to the injury. The non-external factors category captured incidents that did not result from an avoidable external condition but rather from inherent or task-related risks, such as muscle strains during lifting. Finally, the insufficient information category comprised narratives that were too brief, vague, or incomplete to determine any causal factor or hazard condition.

Large Language Models (LLMs), such as GPT-4o, were well-suited for this classification task because of their demonstrated zero-shot and few-shot reasoning capabilities~\cite{llms_are_few_shot_learners}. These models were pre-trained on large and diverse textual corpora, enabling them to generalize to new domains without supervised fine-tuning. Prior research demonstrated that LLMs could infer causal relationships and classify domain-specific narratives using structured prompts~\cite{classificaiton_gpt4o}. In this study, GPT-4o was prompted to simulate a workplace safety inspection, allowing the model to extract contextual information, identify hazards, and classify each narrative according to its preventability, thereby applying occupational safety logic in a zero-shot setting.
\begin{figure}[t] 
    \centering
    \includegraphics[width=1\textwidth]{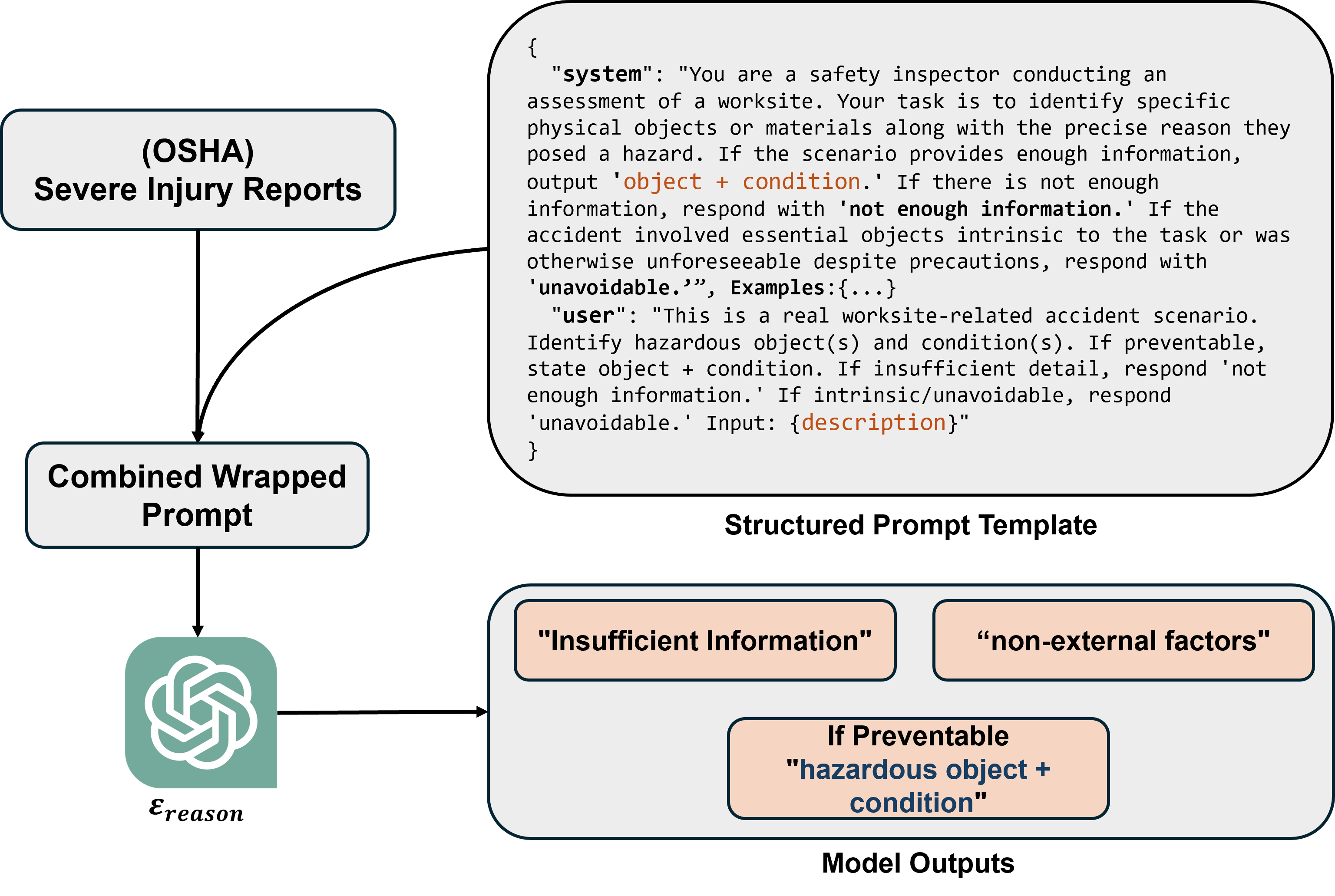}
    \caption{Structured prompt framework for hazard extraction.}
    \label{fig:framework}
\end{figure}
To support this process, a structured prompt was designed to guide the model’s reasoning based on domain-specific logic, as illustrated in Figure~\ref{fig:framework}. The prompt followed an (object+condition) format, which required the model to move beyond simple entity recognition and capture the causal state that defined a hazard. For example, instead of merely identifying the word “ladder,” the model was required to specify a hazardous state such as “ladder placed on uneven surface.” The prompt also incorporated in-context learning elements, including a role definition (“You are a safety inspector”) and several few-shot examples to align the model’s reasoning with occupational safety assessment logic~\cite{llms_are_few_shot_learners}. Negative classifications were also embedded within the prompt to filter out incomplete or irrelevant entries. Narratives with insufficient information were discarded, while those categorized as non-external factors captured incidents not caused by preventable external hazards. This dual-filtering strategy functioned as a high-precision mechanism that isolated a clean and well-defined dataset of identifiable, preventable hazards for subsequent visual generation.

The classification and filtering process was formally expressed as follows. Let $\mathcal{S}$ denote the complete set of $M = 89{,}068$ raw OSHA injury narratives:
\begin{equation}
    \mathcal{S} = \{ s_1, \dots, s_M \} 
\end{equation}
A reasoning function, $\encoder_{\text{reason}}$, was defined to represent the structured prompting pipeline, with GPT-4o~\cite{gpt4} as the underlying model. Applying this function to every narrative $s_i \in \mathcal{S}$ produced a complete, unfiltered output set:
\begin{equation}
    \mathcal{O} = \{ \encoder_{\text{reason}}(s_i) \mid s_i \in \mathcal{S} \} 
\end{equation}
where $|\mathcal{O}| = M = 89{,}068$. The raw output set $\mathcal{O}$ contained all generated responses from the model, distributed across the three classification categories described in Table~\ref{tab:distribution}.

\begin{table}[h!]
  \fontsize{11pt}{13.2pt}\selectfont
  \centering
  \caption{Distribution of GPT-4o Classifications for all OSHA Narratives.}
  \vspace{5pt}
  \label{tab:distribution}
  \begin{tabular}{@{}ll@{}}
  \toprule
  \textbf{Classification Category} & \textbf{Count} \\
  \midrule
  Preventable Hazards (Avoidable) & 35,153 \\
  Non-External Factors & 28,246 \\
  Insufficient Information & 25,669 \\
  \midrule
  Total Entries Processed & 89,068 \\
  \bottomrule
  \end{tabular}
\end{table} 
\vspace{12pt} 
\noindent The final dataset of interest, $\mathcal{H}$, was a filtered subset from $\mathcal{O}$ containing only the valid preventable hazard rationales,
\begin{equation}
\mathcal{H} = \{ o \mid o \in \mathcal{O} \land \operatorname{is\_preventable\_hazard(o)} = \text{True} \}   
\end{equation}
Following this computation, the total number of hazard scenarios used for the clustering stage was $N_{\text{hazards}} = |\mathcal{H}| = 35{,}153$. The set $\mathcal{H}$ represented the collection of structured textual descriptions utilized in the subsequent stage.

\subsection{Semantic Embedding and Clustering of Hazard Rationales}
\label{sec:embedding}

The $N_{\text{hazards}}$ textual rationales in the filtered set $\mathcal{H}$ represented unique incidents, although many shared latent semantic characteristics. To identify these common risk patterns, an unsupervised clustering approach was employed. In this process, the textual descriptions were mapped into a high-dimensional latent space, and the resulting numerical representations were subsequently clustered to group semantically similar hazard scenarios.

For this task, the \texttt{all-MiniLM-L6-v2} model~\cite{sentence_bert,minilm} was used, as it provided efficient and semantically robust sentence-level embeddings suitable for clustering. Each embedding vector, denoted as $\mathbf{z}_i$, captured the conceptual structure of the hazard, allowing for the comparison of similar rationales even when the wording differed; for example, “unsecured ladder on slope” versus “ladder placed on uneven surface.”

This encoding process transformed the textual dataset $\mathcal{H}$ into a numerical dataset $Z$, which contained all embedding vectors:
\begin{equation}
   Z = \{ \mathbf{z}_i \}_{i=1}^{N_{\text{hazards}}} = \{ \encoder_{\text{embed}}(h_i) \mid h_i \in \mathcal{H} \} 
\end{equation}

Here, $\encoder_{\text{embed}}$ represented the embedding function implemented by the sentence transformer model, $h_i$ denoted the $i$-th hazard rationale in $\mathcal{H}$, and $\mathbf{z}_i$ indicated the corresponding embedding vector in $\mathbb{R}^m$, where $m$ was the embedding dimension.

The resulting set $Z$ mapped all hazard rationales into an abstract vector space, where semantically similar incidents were positioned closer together. To identify dominant hazard archetypes, an unsupervised clustering algorithm was then applied to $Z$. This operation partitioned the $N_{\text{hazards}}$ vectors into $k$ distinct semantic clusters, denoted as density-based cluster $C = { C_1, \dots, C_k }$, such that
\begin{equation}
C = \text{Cluster}(Z)
\end{equation}

\subsection{Density-Based Clustering Using HDBSCAN}
\label{sec:cluster}

For the clustering operation $C = \text{Cluster}(Z)$, HDBSCAN~\cite{malzer2020hybrid} was applied. This hierarchical density-based algorithm is well-suited for high-dimensional and noisy data such as sentence embeddings. HDBSCAN constructs a hierarchy of candidate clusters and selects the most stable ones based on their persistence across density levels. 

Two key parameters were defined in this setup. The minimum cluster size was set to $k = 30$, which specified the smallest allowable number of points within a cluster and helped suppress spurious micro-clusters. The minimum samples parameter was set to $n = 10$, which controlled the core point density threshold and governed the algorithm’s robustness to noise and outliers. These parameter values provided a balance between sensitivity to rare patterns and the interpretability and stability of clusters within the latent hazard space.

Each resulting cluster $C_k$ represented a dominant hazard archetype derived from real incident reasoning. For example, one cluster captured scenarios involving "fall from elevation", while another aggregated "slip and trip hazards caused by obstructions". This concept facilitated scalable representation of hazards and targeted synthetic generation.

\begin{figure}[t]
    \centering
    \includegraphics[width=1\textwidth]{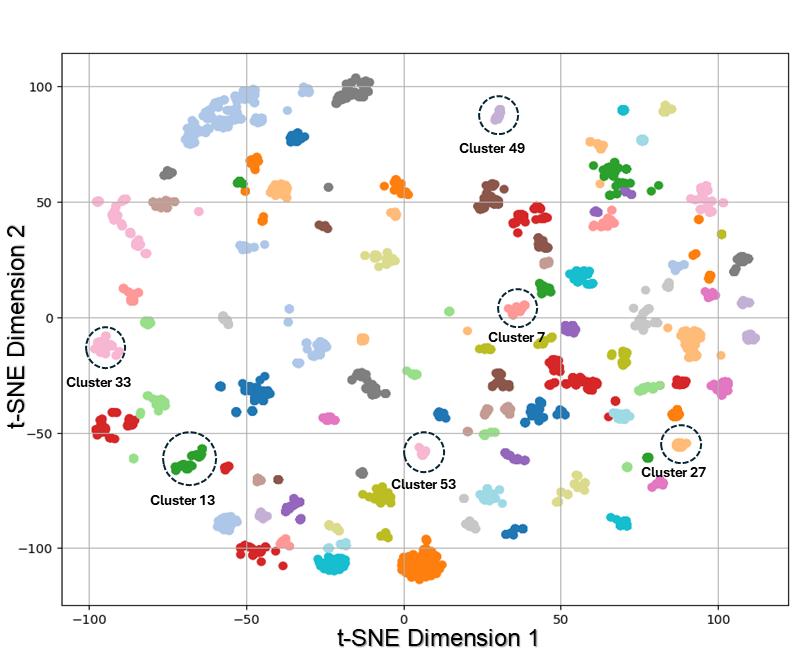}
    \caption{t-SNE visualization of clustered hazard rationales.}
    \label{fig:tsne_clusters}
\end{figure}
 
\subsection{Selection of Representative Hazard Archetypes}
\label{sec:cluster_selection}

The HDBSCAN clustering process successfully transformed the unstructured set of $35,153$ hazard vectors $Z$ into distinct, semantically coherent groups. From these results, the highest-density clusters were manually inspected to identify representative hazard archetypes suitable for image generation. Six dominant clusters were selected, corresponding to the most frequent and clearly defined workplace risks observed in the dataset. These six hazard archetypes, along with their cluster identifiers and item counts, are summarized in Table \ref{tab:hazard_clusters}. The semantic organization of these hazard families, such as walkway obstructions, unguarded openings, cable-related trip hazards, and packaging debris, was visualized within the latent space using a t-SNE projection, as shown in Figure~\ref{fig:tsne_clusters}.
The t-SNE visualization revealed clear semantic separation among the clusters, demonstrating that the embedding and HDBSCAN clustering pipeline effectively captured meaningful patterns within the OSHA narratives. Incidents with similar underlying risk factors were grouped together even when their textual descriptions varied. The annotated clusters highlighted these dominant archetypes, providing an interpretable representation of the most common workplace hazards contained in the dataset.
\begin{table*}[t!]
  \fontsize{11pt}{13.2pt}\selectfont
  \centering
  \caption{Top Hazard Archetypes Identified by Clustering OSHA Narratives}
  \label{tab:hazard_clusters}
  \begin{tabular}{@{} l c p{6.8cm} p{6.5cm} @{}}
  \toprule
  \textbf{ID} & \textbf{Samples} & \textbf{Hazard Archetype} & \textbf{Example Rationale} \\
  \midrule
  7  & 197 & Walkway Obstruction Hazards & Chair leg extended into walkway \\
  53 & 120 & Cable and Cord Trip Hazards & Electrical cord improperly secured or placed \\
  13 & 110 & Unguarded Floor Opening Hazards & Existing floor opening without proper guardrails \\
  49 & 97  & Trip Hazards from Plastic Packaging Debris & Plastic strapping left on workplace floor \\
  33 & 76  & Unguarded Scaffolding Hazards & Scaffold without proper guardrails \\
  27 & 65  & Wet Floor and Liquid Spill Hazards & Water spilled on floor \\
  \bottomrule
  \end{tabular}
\end{table*}
While the generation framework was designed to synthesize images for all identified archetypes, the experimental evaluation in this study was conducted on a single, challenging benchmark: Cluster 49, which contained trip hazards from plastic packaging debris. This cluster was specifically selected because it represented a high-frequency yet subtle hazard type. Unlike large-scale hazards such as unguarded scaffolds, small pieces of plastic strapping on the floor required high compositional accuracy and contextual awareness from the generative model to be depicted as a clear hazard. This characteristic made Cluster 49 an ideal and rigorous test case for validating the fidelity of our scene graph-guided generation pipeline and the discriminative power of our proposed evaluation metrics.

\subsection{Scene Graph-Based Scenario Modeling}\label{sec:vqa}

Unlike purely text-driven or end-to-end image generation pipelines, the proposed framework incorporated an explicit intermediate representation in the form of a scene graph. This graph-based inference functioned as a critical intermediary, translating narrative hazard descriptions into structured visual cues. Formally, each scene graph consisted of a set of nodes representing physical objects, such as "metal ladder" and "spill". Additionally, each scene graph contained directed edges that captured relational assertions, such as 'on top of' or 'blocking'. Attributes such as material type or physical state were embedded at the node level. This structured encoding enabled the disentanglement of high-level safety semantics from low-level details. In this way, consistency between the hazard logic and the final image was ensured.

Once a representative hazard cluster $C_j \in C$, specifically Cluster 49, had been selected, its corresponding set of textual rationales $\mathcal{H}_j \subset \mathcal{H}$ was retrieved. Each rationale $h_i \in \mathcal{H}j$ was then passed through a dedicated scene graph encoder, denoted as $\encoder_{\text{SgGen}}$. For this encoding step, the LLaMA 3.3 model with 70 billion parameters was utilized~\cite{dubey2024llama}.

\begin{figure}[t]
    \centering
    \includegraphics[width=.8\textwidth]{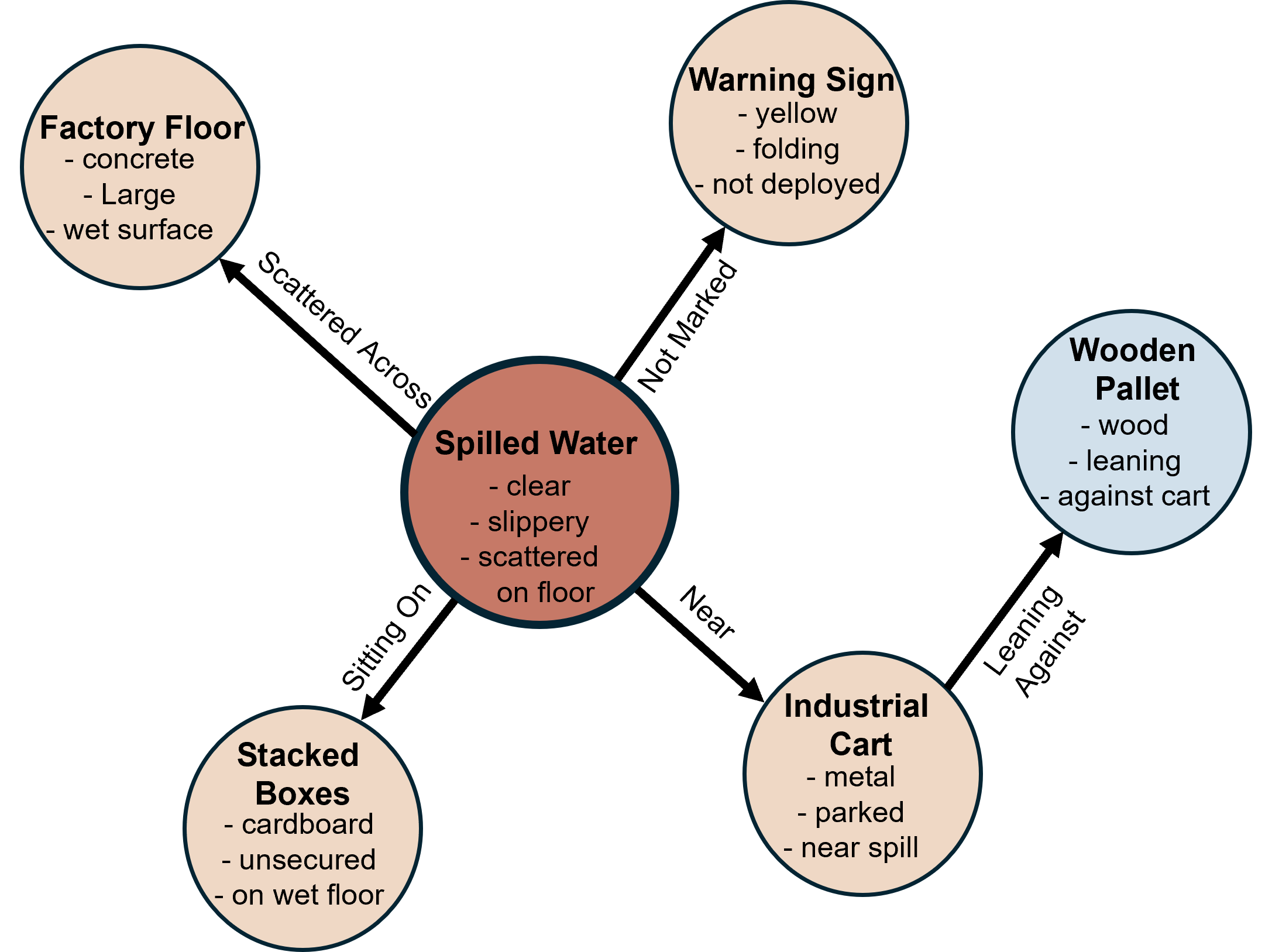}
    \caption{Example scene graph for a slip hazard.}
    \label{fig:scenegraph}
\end{figure}

Using $\encoder_{\text{SgGen}}$, the task of contextual elaboration was performed. The model was prompted to construct a plausible and coherent workplace scenario in which the given hazard $h_i$ could logically occur. The encoder identified key physical objects, such as platforms or ladders, to serve as graph nodes $\mathcal{V}_i$ and assigned each node contextual attributes, such as material type or orientation. These nodes were then organized through directed edges $\mathcal{R}_i$ representing spatial or causal relationships, such as "leaning against" or "placed on". This generated configuration captured the essential risk structure in an interpretable and reusable format. The final output from this step was a complete scene graph $\mathcal{G}_i = (\mathcal{V}_i, \mathcal{R}_i)$, a transformation formally expressed as,
\begin{equation}
   \mathcal{G} = \{ \encoder_{\text{SgGen}}(h_i) \mid h_i \in \mathcal{H} \}
\end{equation}
A representative example of a scene graph $\mathcal{G}_i$ for a slip hazard from Cluster 27, one of the major hazard classes, is shown in Figure~\ref{fig:scenegraph}. The graph defined a "Factory Floor" node with attributes "concrete" and "wet surface," and identified the primary hazard node, "Spilled Water," with attributes "clear" and "slippery." Contextual objects were also included, such as a "Warning Sign" (node) that was "not deployed," "Stacked Boxes" (node) placed "on wet floor," and an "Industrial Cart" (node) "parked near spill." By encoding objects, attributes, and their relationships, the scene graph preserved the causal and spatial logic of the unsafe scenario in a structured and interpretable format, serving as a reusable intermediate layer for both image synthesis and evaluation.

\subsection{Vision Language Evaluation and Graph-Aligned Prompt Generation.}\label{sec:vqa_eval}

\begin{figure}[t]
\centering
\includegraphics[width=1\textwidth]{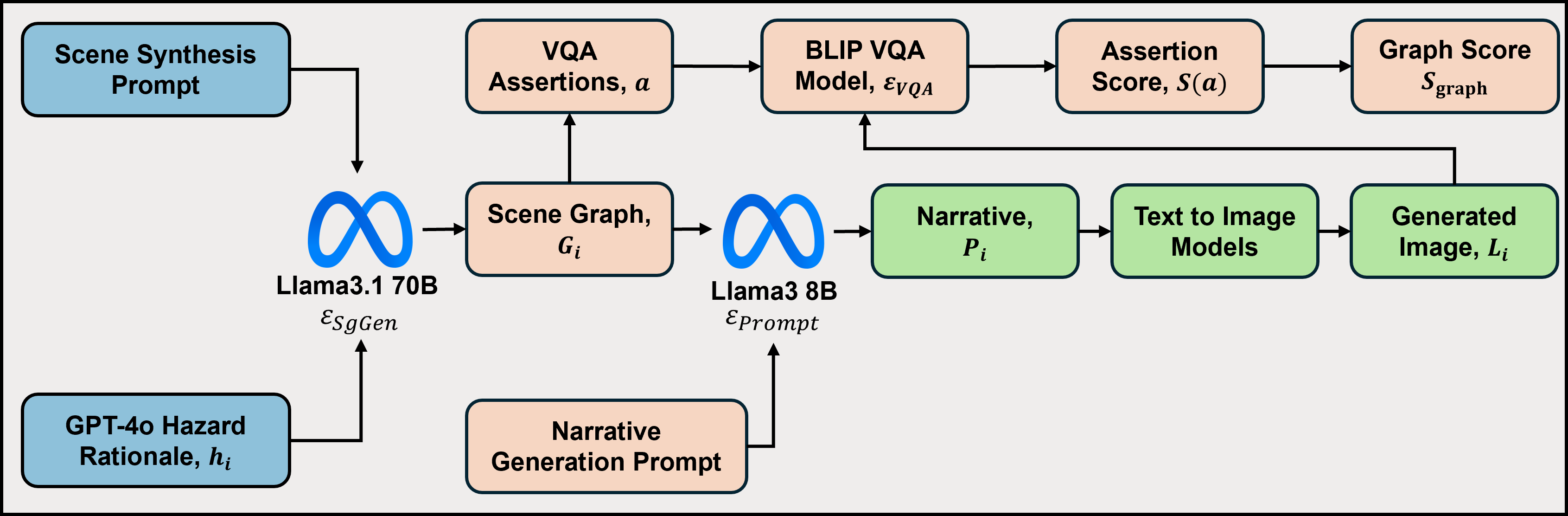}
\caption{Pipeline for scene generation and evaluation.}
\label{fig:method_flow}
\end{figure}

Once the set of scene graphs $\mathcal{G}$ had been generated, each graph $\mathcal{G}_i \in \mathcal{G}$ served a dual purpose: it guided the image synthesis process and simultaneously provided the ground-truth benchmark for evaluation. Figure~\ref{fig:method_flow} illustrates this workflow. Starting from GPT-4o derived hazard rationales, structured scene graphs were constructed to encode entities, attributes, and relationships. Each graph was then transformed in two ways: (a) into natural language prompts for text-to-image synthesis and (b) into declarative assertions for evaluation. Unlike traditional VQA tasks that rely on open-ended questions, the evaluation framework treated each graph as a structured verification task in which nodes (objects and attributes) and edges (relationships) were systematically translated into assertions. A pretrained vision language model subsequently verified whether each assertion was visually supported in the generated image. The results were then aggregated into a structured compliance score. This unified pipeline ensured that both generation and evaluation remained grounded in the same semantic representation.

To achieve this, the generative prompt was created by passing each graph $\mathcal{G}_i$ to a text-generation encoder, $\encoder_{\text{Prompt}}$, powered by the LLaMA 3.1 7B model. This function generated a short, natural-language narrative summering the hazard scenario:
\begin{equation}
p_i = \encoder_{\text{Prompt}}(\mathcal{G}_i)
\end{equation}
The generated text prompt $p_i$ was then provided to the Stable Diffusion model to synthesize the corresponding image $I_i$.

For the evaluation step, each scene graph $\mathcal{G}_i = (\mathcal{V}_i, \mathcal{R}_i)$ was converted into a set of verifiable assertions, $A_i$, to create the benchmark. This set was composed of attribute-level assertions for each node $v \in \mathcal{V}_i$, such as “The platform is metal,” and relational assertions for each edge $e \in \mathcal{R}_i$, such as “The hard hat is below the platform edge.”

Each assertion $a \in A_i$ was then evaluated against the generated image $I_i$ using a pretrained vision–language model such as BLIP~\cite{blip}, denoted as the VQA encoder $\encoder_{\text{VQA}}$. For this task, BLIP2~\cite{blip2} was used. The encoder returned a soft confidence score $s(a) \in [0, 1]$, representing the likelihood that the assertion was visually satisfied in the image:
\begin{equation}
   s(a) = \encoder_{\text{VQA}}(I_i, a)
\end{equation}
These individual assertion scores formed the basis of the structured, graph-aligned image evaluation metric described below.

To obtain a holistic measure of how accurately the generated image reflected the corresponding scene graph, the individual assertion scores $s(a)$ were aggregated into a single, graph-level compliance score. For any given scene graph $\mathcal{G} = (\mathcal{V}, \mathcal{R})$, this process was defined as follows. Each node $v_j \in \mathcal{V}$ had a set of attribute-level assertions $\mathcal{A}(v_j)$, where $\mathcal{A}(v_j)$ denotes the set of all attribute-level assertions associated with node $v_j$. Each edge $r_k \in \mathcal{R}$ corresponded to a single relational assertion $a_{r_k}$. The average score for each node, $S_{v_j}$, and the score for each edge, $S_{r_k}$, were computed as:
\begin{align}
    S_{v_j} &= \frac{1}{|\mathcal{A}(v_j)|} \sum_{a \in \mathcal{A}(v_j)} s(a), \quad \text{and} \\
    S_{r_k} &= s(a_{r_k})
\end{align}
where $s(a) = \encoder_{\text{VQA}}(I, a)$ represented the confidence score produced by the VQA encoder for assertion $a$ in image $I$.

To emphasize critical elements of the hazard, a key node $v_h \in \mathcal{V}$ corresponding to the primary hazard object was assigned a higher weight. Here, $v_h$ denotes the primary hazard node within the scene graph, $\lambda > 1$ represents the node-level weight multiplier applied to $v_h$, and $\gamma > 1$ represents the edge-level weight multiplier applied to all edges incident to $v_h$. The weights $w_{v_j}$ and $w_{r_k}$ were then defined as:
\begin{align}
w_{v_j} &= 
\begin{cases}
\lambda, & \text{if } v_j = v_h \\
1,       & \text{otherwise}
\end{cases} \\
w_{r_k} &= 
\begin{cases}
\gamma, & \text{if } r_k \text{ is incident to } v_h \\
1,      & \text{otherwise}
\end{cases}.
\end{align}

The overall graph-level compliance score, $S_{\text{graph}}$, was then calculated as a weighted average over all nodes and edges:
\begin{equation}    
S_{\text{graph}} = \frac{
\sum_{v_j \in \mathcal{V}} w_{v_j} S_{v_j} + \sum_{r_k \in \mathcal{R}} w_{r_k} S_{r_k}
}{
\sum_{v_j \in \mathcal{V}} w_{v_j} + \sum_{r_k \in \mathcal{R}} w_{r_k}
}\label{eq:S_graph}
\end{equation}
Here, $S_{\text{graph}} \in [0, 1]$ denotes the graph-level compliance score, which quantifies how closely the generated image preserved the semantic and relational structure encoded in the original scene graph. This formulation ensured that the entire graph contributed to the evaluation while assigning greater importance to the components most critical for hazard interpretation. The resulting scalar score $S_{\text{graph}} \in [0,1]$ reflected the extent to which the generated image preserved the semantic and relational integrity of the original hazard scenario. This score was subsequently used to filter low-fidelity scenes, rank generated examples, and selectively regenerate those that failed to meet a desired level of semantic alignment.

\section{Experimental Results}\label{sec:results}

\noindent To evaluate the quality and semantic fidelity of the generated hazard scenes, a comprehensive analysis was conducted using three complementary evaluation methods:
(1) the CLIP score, measuring alignment between the generated image and the textual prompt;
(2) the BLIP matching score, which assessed image–language consistency using a pretrained vision-language model; and
(3) the proposed graph-aligned VQA-based evaluation, which directly verified the structural and contextual accuracy of the generated scenes against the corresponding input scene graphs.
All scores were normalized to the range $[0, 1]$, thereby ensuring comparability across evaluation methods.

\subsection{Evaluation of State-of-the-Art Generative Models}
\begin{figure} [t!] 
    \centering
    \includegraphics[width=1\textwidth]{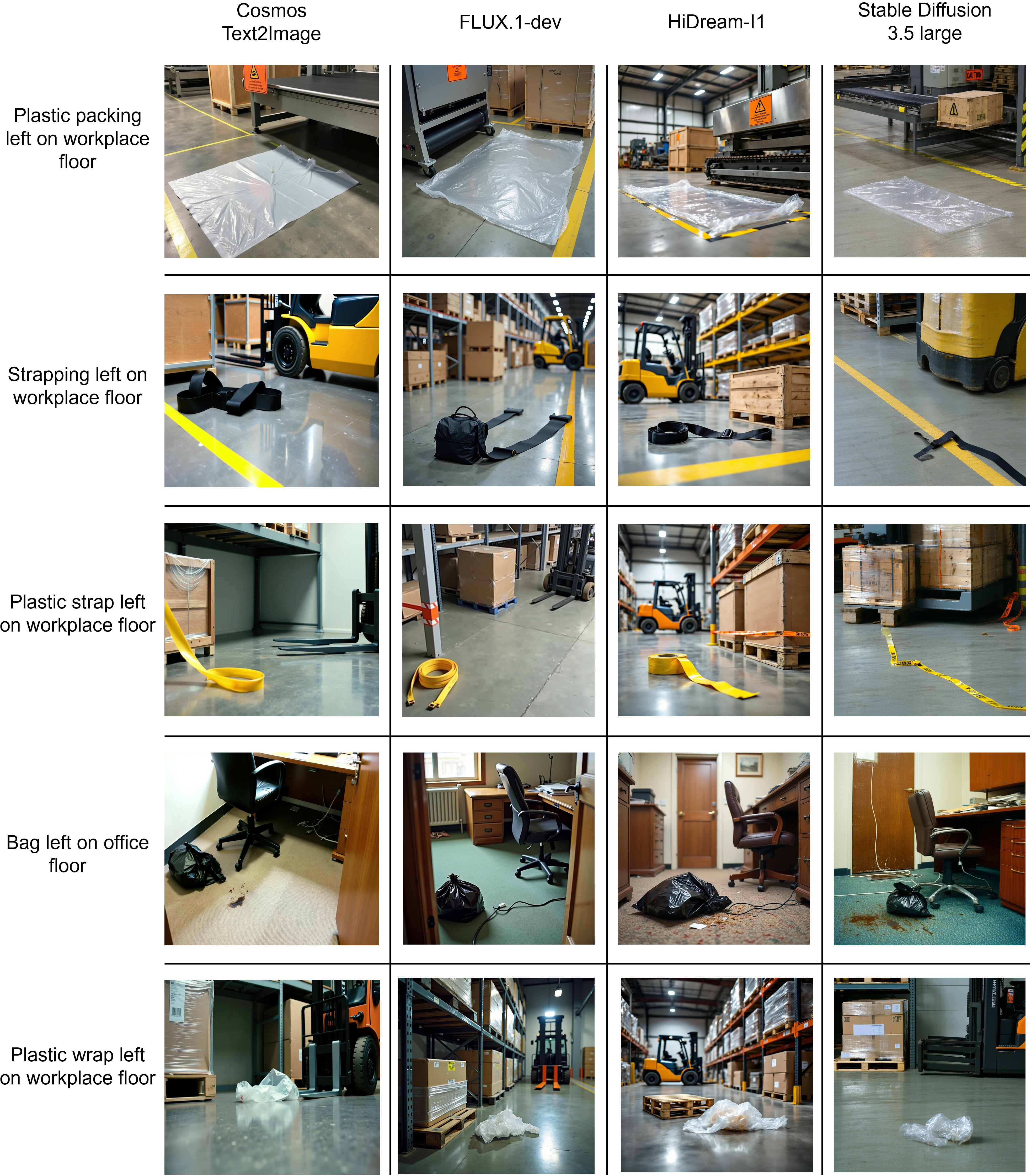}
    \caption{Outputs from four generative models across hazard prompts.}
    \label{fig:generated_outputs_grid}
\end{figure}

In this experiment, the performance of four leading text-to-image models was compared on a representative set of hazard scenarios: Stable Diffusion 3.5 Large (SD-3.5L)~\cite{stable_diffusion}, FLUX.1-dev (FLUX)~\cite{flux}, NVIDIA Cosmos (Cosmos) Text2Image~\cite{cosmos}, and HiDream-I1(HiDream)~\cite{hidream}. To provide context for their performance, the parameter scales of these models are summarized in Table~\ref{tab:model_params}. 
\begin{table}[ht]
\fontsize{11pt}{13.2pt}\selectfont
\centering
\caption{Parameter counts of the generative models evaluated for this study.}
\vspace{5pt}
\label{tab:model_params}
\begin{tabular}{l r}
\toprule
\textbf{Model} & \textbf{Parameters} \\
\midrule
SD 3.5 large  & 8.1B \\
FLUX.1-dev    & 12B  \\
NVIDIA Cosmos & 14B  \\
HiDream-I1    & 17B  \\
\bottomrule
\end{tabular}
\end{table}
As shown in Figure~\ref{fig:generated_outputs_grid}, the four generative models produced noticeably different renderings of the same hazard prompts. Although all models were capable of synthesizing plausible warehouse or office environments, their ability to depict specific hazards varied considerably. For instance, HiDream-I1 integrated plastic packaging debris more naturally into the floor environment, whereas Stable Diffusion often failed to represent the hazard clearly or generated unrealistic overlays. These qualitative differences underscored the need for evaluation methods that move beyond general text–image alignment and toward fine-grained compositional fidelity.
The quantitative evaluation was conducted to assess how faithfully each generated image reflected its corresponding hazard description, using metrics that approached the comparison in different ways. For CLIP and BLIP, the assessment was performed within a shared multimodal embedding space. A text encoder, $\encoder_{\text{CLIP text}}$ mapped the summary prompt ($p$) to a feature vector, $\mathbf{v}_p = \encoder_{\text{CLIP text}}(p)$, while a vision encoder, $\encoder_{\text{CLIP image}}$, mapped the generated image, $I$, to a vector of the same dimension, $\mathbf{v}_I = \encoder_{\text{CLIP image}}(I)$. The alignment between the two modalities was then measured by the similarity between these vectors. The CLIP score, for example, corresponded to the cosine similarity between $\mathbf{v}_p$ and $\mathbf{v}_I$, as defined in Equation~\ref{eq:cosine_sim}. The BLIP matching score followed a similar principle but employed a cross-modal attention mechanism to evaluate image and text consistency. In contrast, the proposed VQA Graph Score conducted a more direct and structured evaluation. Rather than relying on holistic vector similarity, it systematically queried each generated image to verify whether the discrete assertions, objects, attributes, and relationships that were derived from the original scene graph were accurately rendered. This approach directly measured compositional fidelity and provided a fine-grained assessment of semantic and spatial alignment.

For the quantitative assessment, the mean and standard deviation of the scores for each model across the three evaluation metrics were reported in Table~\ref{tab:model_comparison}. The results revealed a critical insight: both the CLIP Score and BLIP Score were largely insensitive to the nuanced differences among the models. All four models achieved nearly identical and saturated scores, suggesting that they successfully captured the high-level topic of the prompt but failed to provide discriminative feedback on compositional accuracy.
\begin{table}[ht]
\fontsize{11pt}{13.2pt}\selectfont
\centering
\caption{Evaluation of scene generation quality across image generation models using three scoring methods.}
\vspace{5pt}
\setlength{\tabcolsep}{3pt}\renewcommand{\arraystretch}{1.0}
\begin{tabular}{lrrr}
\toprule
\textbf{Model} & \textbf{CLIP Score} & \textbf{BLIP Score} & \textbf{VQA Graph Score} \\
\midrule
Stable Diffusion    & $0.31 \pm 0.00$ & $0.89 \pm 0.02$ & $0.35 \pm 0.01$ \\
FLUX                & $0.30 \pm 0.00$ & $0.94 \pm 0.01$ & $0.46 \pm 0.01$ \\
NVIDIA Cosmos       & $0.29 \pm 0.00$ & $0.94 \pm 0.01$ & $0.50 \pm 0.01$ \\
HiDream             & $0.31 \pm 0.00$ & $0.93 \pm 0.01$ & $0.52 \pm 0.01$ \\
\bottomrule
\end{tabular}
\label{tab:model_comparison}
\end{table}

To further illustrate the quantitative findings and provide a qualitative comparison, a grid of example outputs from each generative model for several hazard archetypes was presented in Figure~\ref{fig:generated_outputs_grid}. This visualization enabled a direct visual inspection of how each model interpreted and rendered the same hazard prompt. For example, the varying quality and coherence in depicting scenarios such as "plastic packing left on workplace floor" or "strapping left on workplace floor" could be observed across the different models. Although all models were capable of generating plausible warehouse environments, their success in accurately and realistically positioning the specific hazardous object (such as a plastic sheet, black strapping, yellow strap, or plastic bag) varied considerably. Some models demonstrated better integration of the hazard into the scene, whereas others appeared to merely overlay objects or produced unrealistic material properties, such as the transparency of plastic wrap. These visual discrepancies highlighted the limitations of relying solely on general-purpose metrics and reinforced the need for a fine-grained and compositional evaluation approach, such as the proposed VQA Graph Score, to more accurately assess the fidelity of generated hazard scenarios.

In contrast, the proposed VQA Graph Score provided a clear and decisive performance ranking that was not captured by the other metrics. This score revealed a wide dynamic range across the evaluated models, ranging from 0.35 for Stable Diffusion to a top score of 0.52 for HiDream. According to this metric, HiDream (0.52 ± 0.01) demonstrated the highest fine-grained alignment, most accurately rendering the specific features and relationships defined in the source scene graph. The performance hierarchy established by the VQA Graph Score was strongly correlated with the models’ parameter counts: HiDream (17B) outperformed NVIDIA Cosmos (14B), which in turn surpassed FLUX (12B) and Stable Diffusion (8.1B). These results suggested a direct relationship between model scale and compositional understanding. In comparison, the rankings provided by the CLIP and BLIP scores appeared more arbitrary; for example, CLIP assigned nearly identical scores to the largest and smallest models. Overall, only the VQA Graph Score provided a consistent and interpretable measure of each model’s ability to adhere to complex and structured prompts.

\begin{figure}[H]
    \centering
    \includegraphics[width=1\textwidth]{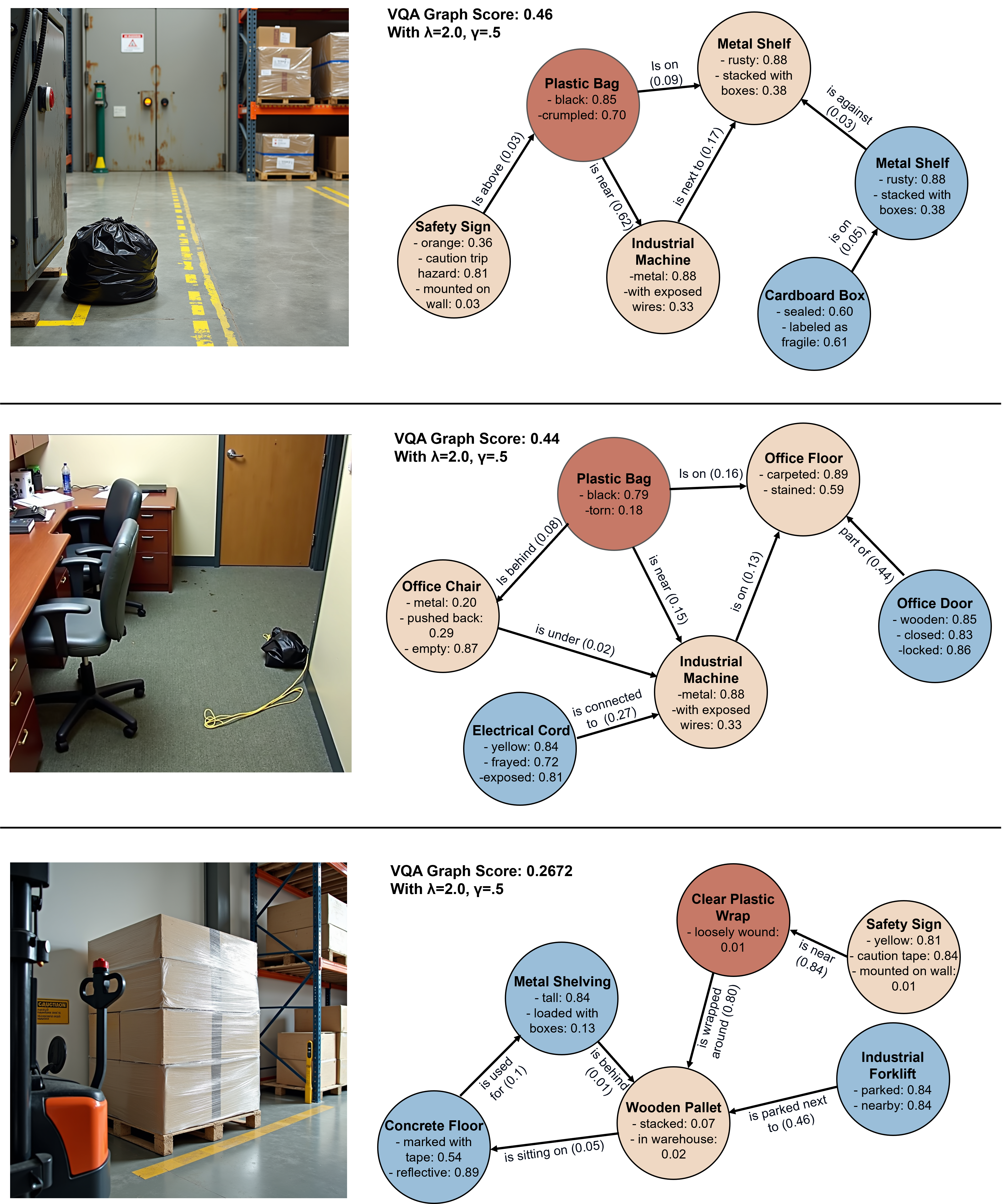}
    \caption{Generated hazard scenarios with scene graphs and VQA Graph Scores.}
    \label{fig:qualitative_results}
\end{figure}
To provide a qualitative illustration of these findings, Figure~\ref{fig:qualitative_results} presented generated hazard scenes alongside their corresponding scene graphs and VQA Graph Scores. The figure highlighted the sensitivity of the metric to compositional accuracy. The bottom panel, which was intended to represent a hazard caused by loosely wound plastic wrap, received a low score of 0.27. Although the overall warehouse environment appeared plausible, the hazardous condition itself was ambiguous and poorly rendered. In contrast, the first (0.46) and second (0.44) panels represent clearer and more well-defined trip hazards, including a plastic bag in a walkway and a stray electrical cord in an office, which aligned more consistently with their respective scene graphs. These examples demonstrate that the VQA Graph Score effectively captures whether the critical, risk-defining elements of a hazard are consistently represented, providing a more detailed assessment than holistic similarity metrics such as CLIP or BLIP.

\subsection{Evaluation of Metric Sensitivity Using Score Distributions and Information Entropy}

To test the sensitivity and reliability of the evaluation metrics, a negative control experiment was conducted. Two sets of intentionally mismatched prompt–image pairs were created: (1) an out-of-domain set, in which the hazard prompts were paired with random images from the Flickr30k~\cite{flicker} dataset, and (2) a more challenging in-domain set, in which the same prompts were paired with randomly shuffled images from the generated dataset. Comparing these mismatched pairs with the correct prompt–image pairs helped evaluate whether each metric could reliably detect misalignment.

\begin{figure}[H]
    \centering
    \includegraphics[width=1\textwidth]{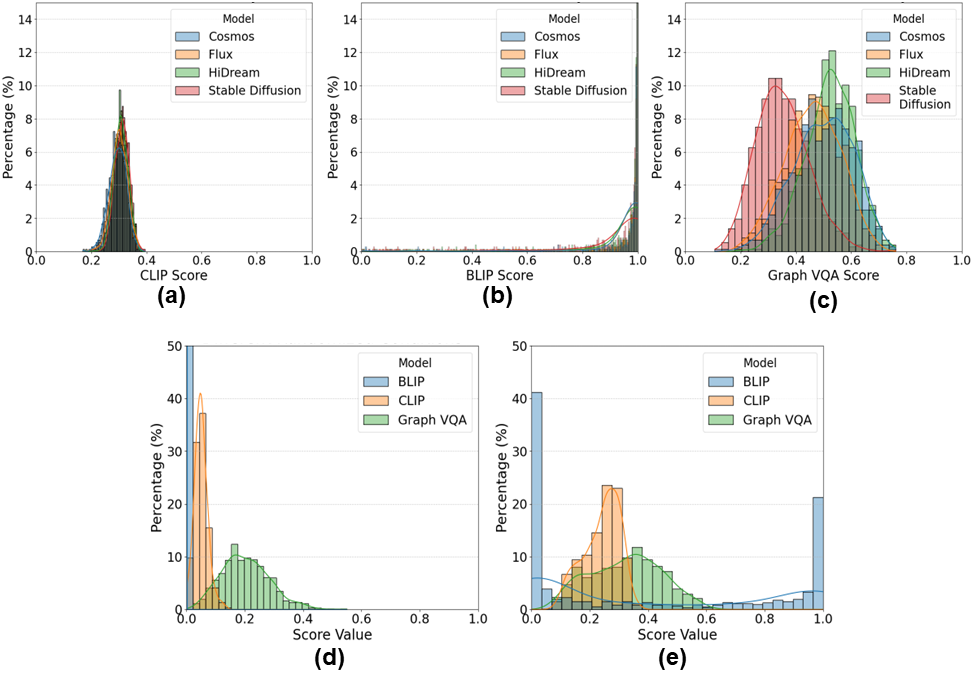}
    \caption{
Score distribution comparisons across evaluation metrics and models.
(a) Distribution of CLIP scores for all generative models.
(b) Distribution of BLIP scores for the same models.
(c) Distribution of Graph VQA scores showing clearer separation and sensitivity among models.
(d) Comparison of CLIP, BLIP, and Graph VQA score distributions under different randomized conditions.
(e) Comparison of CLIP, BLIP, and Graph VQA scores under identical randomized conditions within the same distribution.
The results highlight the higher discriminative capability of the Graph VQA metric compared with embedding-based scores.
}
    \label{fig:score_distribution}
\end{figure}

The score distributions presented in Figure~\ref{fig:score_distribution} revealed clear differences in how the evaluation metrics responded to matched and mismatched prompt–image pairs. For correctly matched pairs (top row), the CLIP scores had low dispersion, whereas the BLIP scores were heavily skewed toward the maximum value of 1.0. 

The VQA Graph Score, however, showed a wider and more normally distributed range of values, suggesting it could capture subtle variations even among valid images. When mismatched pairs were used (bottom row), the differences became clearer. BLIP scores dropped sharply to near zero for random Flickr images, but were inconsistent for shuffled in-domain examples. CLIP scores remained broad and overlapped with the matched distributions, making it difficult to distinguish between correct and incorrect pairs. The VQA Graph Score, on the other hand, consistently shifted toward lower values in both cases, indicating that it was more sensitive and effective at identifying compositional errors than the embedding-based metrics.
\vspace{100pt}
Table~\ref{tab:metrics_entropy_restyled} presented the Shannon entropy values for the score distributions of the four generative models, measuring the informativeness of three different evaluation metrics: BLIP, CLIP, and the proposed VQA Graph Score. For all evaluated models, the VQA Graph Score consistently provided the highest entropy, for example, 3.98 bits for FLUX and 3.82 bits for NVIDIA Cosmos, followed by the CLIP score, whereas the BLIP score exhibited the lowest entropy, such as, 1.19 bits for FLUX and 1.14 bits for NVIDIA Cosmos. From an information-theoretic perspective, a higher entropy indicates a less predictable and more varied distribution of scores. This result suggests that the VQA Graph Score provides a more descriptive and fine-grained assessment, conveying a greater amount of information about variations in the generated images. Conversely, the low entropy of the BLIP score implied that its outputs were highly concentrated and saturated, offering limited discriminative power and less informational content regarding the compositional accuracy of the images.
\begin{table}[H]
    \fontsize{11pt}{13.2pt}\selectfont
    \caption{Shannon Entropy for model scores and compliance.}
    \vspace{5pt}
    \centering
    \begin{tabular}{llr} 
    \toprule 
    \multicolumn{3}{c}{\textbf{Score Distribution's Entropy ($\uparrow$)}} \\
    \\ 
    \textbf{Model} & \textbf{Score Type} & \textbf{Shannon Entropy (bits)} \\
    \midrule 
    \multirow{3}{*}{NVIDIA Cosmos} & BLIP & 1.14 \\
     & CLIP & 3.56 \\
     & VQA Graph & \textbf{3.82} \\
    \midrule 
    \multirow{3}{*}{Flux} & BLIP & 1.19 \\
     & CLIP & 3.60 \\
     & VQA Graph & \textbf{3.98} \\
    \midrule
    \multirow{3}{*}{HiDream} & BLIP & 1.49 \\
     & CLIP & 3.41 \\
     & VQA Graph & \textbf{3.70} \\
    \midrule
    \multirow{3}{*}{Stable Diffusion} & BLIP & 1.99 \\
     & CLIP & 3.54 \\
     & VQA Graph & \textbf{3.58} \\
    \bottomrule 
    \end{tabular}
    \label{tab:metrics_entropy_restyled} 
\end{table}

Table~\ref{tab:metrics_entropy_shuffled} extended this analysis by examining the entropy of the metrics when prompts were intentionally mismatched with images. In the case of the out-of-domain shuffle using random Flickr images, the CLIP score and the VQA Graph Score produced nearly identical entropy values (3.71 bits and 3.64 bits, respectively), indicating that the VQA Graph Score was not inherently superior when evaluating completely irrelevant image–prompt pairs. However, a critical distinction emerged in the more challenging in-domain shuffle, which used stylistically similar images from the same generated distribution. In this scenario, the VQA Graph Score achieved a higher entropy value (4.11 bits), surpassing the CLIP score (3.98 bits). This higher informational content demonstrated that, for the nuanced task of identifying compositional errors among plausible but incorrect images, the VQA Graph Score functioned as the more sensitive and discriminative metric.

\begin{table}[H]
    \fontsize{11pt}{13.2pt}\selectfont
    \caption{Shannon Entropy for Shuffled Scores}
    \centering
    \vspace{5pt}
    \setlength{\tabcolsep}{3pt}\renewcommand{\arraystretch}{1.0}
    \begin{tabular}{llr} 
    \toprule 
    \multicolumn{3}{c}{\textbf{Shuffled Score Distribution's Entropy ($\uparrow$)}} \\
    \\ 
    \textbf{Shuffle Source} & \textbf{Score Type} & \textbf{Shannon Entropy (bits)} \\
    \midrule 
    \multirow{3}{*}{Flickr (Random)} & BLIP & 0.02 \\
     & CLIP & \textbf{3.71} \\
     & VQA Graph & 3.64 \\
    \midrule 
    \multirow{3}{*}{Same Distribution (Random)} & BLIP & 2.90 \\
     & CLIP & 3.98 \\
     & VQA Graph & \textbf{4.11} \\
    \bottomrule 
    \end{tabular}
    \label{tab:metrics_entropy_shuffled} 
\end{table}

\section{Limitations}\label{sec:limitations}

\noindent \textit{Error Propagation Across the Generation Pipeline:} The sequential design of the framework made it susceptible to error accumulation across its stages. The quality of the final generated image was dependent on the accuracy of each preceding step. If the language model failed to construct an accurate or coherent scene graph from the input hazard rationale, even an advanced diffusion model produced a flawed or irrelevant scene. This interdependence created a cascading effect in which weaknesses at any stage, from narrative interpretation to image synthesis, degraded the overall quality of the generated hazard scenario.

\vspace{6pt}
\noindent \textit{Limitations of the VQA Model in Evaluation Accuracy:} The reliability of the VQA Graph Score was inherently constrained by the perceptual and reasoning capabilities of the vision-language model used for evaluation. The accuracy of the score depended on the model’s ability to identify objects, interpret their attributes, and verify spatial relationships within the generated image. Any shortcomings or biases in the vision-language model, such as difficulty in counting, recognizing subtle physical states like “wet” or understanding spatial relations such as “below” or “beside” were directly reflected in the resulting evaluation scores. Consequently, the fidelity of the proposed metric was bounded by the current level of vision-language understanding.

\vspace{6pt}
\noindent \textit{Semantic Control in Scene Elaboration:} Using a large language model to elaborate hazard scenarios added contextual richness but also introduced potential semantic inconsistencies. The model occasionally generated plausible yet contextually inappropriate elements that undermined the depiction of the intended hazard. For instance, when describing a slip hazard, the language model sometimes included a “wet floor” warning sign, which neutralized the risk being represented. Although careful prompt engineering mitigated this issue to some extent, achieving precise semantic control to ensure that the generated scene consistently reflected an unsafe condition rather than a managed one remained a significant challenge.

\section{Conclusions}\label{sec:conclusion}

\noindent This study was designed to address the shortage of visual data needed to train intelligent workplace hazard-detection systems. By linking textual accident narratives with visual scene understanding, a scene-graph-guided generative AI framework was proposed. A method capable of producing coherent and context-aware visual hazard scenarios grounded in  OSHA reports.

The research introduced a structured methodology that integrates LLM reasoning, semantic clustering, and scene graph modeling with diffusion-based image synthesis. To create a robust evaluation pipeline for this method, a novel VQA based graph score was designed. The VQA Graph Score can assess the semantic and compositional accuracy of the generated images, outperforming conventional measures such as CLIP or BLIP. The experimental results showed that the VQA Graph metric captured more meaningful differences, offering an average of 2.32 more bits of information than BLIP and an average of 0.24 more bits of information than CLIP.. Additionally, across generative models, the VQA Graph Score showed a strong correlation with model scale and compositional fidelity. Among the evaluated models, HiDream had the most robust scene synthesis performance, achieving the highest VQA Graph Score of 0.52.

The approach suggests a broader shift toward proactive safety modeling, where AI systems learn from simulated, hypothetical hazard scenarios. The graph-aligned evaluated data produced through this framework can serve as a training resource for next-generation hazard detection models, enabling them to recognize rare or unseen safety risks with improved generalization and robustness. Such data can also support simulation-based training, predictive risk analysis, and cross-domain data augmentation for several industrial applications.

\section*{Credit Authorship Contribution Statement}\label{sec:credit}
\noindent Sanjay Acharjee contributed to conceptualization, methodology, formal analysis, visualization, validation, and writing original draft. Abir Khan Ratul contributed to conceptualization, methodology, validation, writing original draft, and writing review and editing. Diego Patiño contributed to conceptualization, methodology, supervision, and writing review and editing. Md Nazmus Sakib, contributed to conceptualization, methodology, supervision, and writing review and editing.

\section*{Declaration of Competing Interest}\label{sec:dec_com}
\noindent The authors declare that they have no known competing financial interests or personal relationships that could have appeared to influence the work reported in this paper.

\section*{Acknowledgments}\label{sec:ack}
\noindent The authors would like to express their sincere gratitude to the Occupational Safety and Health Administration (OSHA) for providing access to the Severe Injury Reports dataset, which served as the foundation for this study.
\section*{Data Availability}\label{sec:data}
\noindent The data supporting the findings of this study are available from the corresponding author upon reasonable request. 
\section*{Deceleration of Generative AI and AI-Assisted Technologies in the Manuscript Preparation Process}\label{sec:genai}
\noindent During the preparation of this work, the authors used Grammarly and GPT-4o to check for plagiarism and grammar. After using these tools, the authors carefully reviewed and edited all content and take full responsibility for the final version of the manuscript.

\nocite{IEEEexample:BSTcontrol}
\bibliographystyle{IEEEtran} 
\bibliography{cas-refs3}

\end{document}